# Benchmarking ChatGPT and DeepSeek in April 2025: A Novel Dual-Perspective Sentiment Analysis Using Lexicon-Based and Deep Learning Approaches

Maryam Mahdi Alhusseini [a*] (Member, IEEE), Mohammad-Reza Feizi-Derakhshi [b]

[a] Information and Communication Technology Department, Middle Technical University, Baghdad, Iraq,
[b] Computerised Intelligence Systems Laboratory, Department of Computer Engineering, University of Tabriz, Tabriz, Iran

*Corresponding Author:* *mariammahdi@mtu.edu.iq

***Abstract***

This study presents a novel dual-perspective approach to analyzing user reviews for ChatGPT and DeepSeek on the Google Play Store, integrating lexicon-based sentiment analysis (TextBlob) with deep learning classification models, including Convolutional Neural Networks (CNN) and Bidirectional Long Short-Term Memory (Bi-LSTM) Networks. Unlike prior research, which focuses on either lexicon-based strategies or predictive deep learning models in isolation, this study conducts an extensive investigation into user satisfaction with Large Language Model (LLM) -based applications. A Dataset of 4,000 authentic user reviews was collected, which were carefully pre-processed and subjected to oversampling to achieve balanced classes. The balanced test set of 1,700 reviews was used for model testing. Results from the experiments reveal that ChatGPT received significantly more positive sentiment than DeepSeek. Furthermore, deep learning–based classification demonstrated superior performance over lexicon analysis, with CNN outperforming Bi-LSTM by achieving 96.41% accuracy and near-perfect classification of negative reviews, alongside high F1-scores for neutral and positive sentiments. This research sets a new methodological standard for measuring sentiment in LLM-based applications and provides practical insights for developers and researchers seeking to improve user-centric AI system design.

## 1 Introduction and Background

The development of large language models (LLMs) has recently seen significant progress, such as GPT-3 [1], PaLM [2], LLaMA [3, 4], ChatGPT [5, 4, 6], and DeepSeek [7, 4, 6]. The rapid advancement of Artificial Intelligence (AI) enhanced language models has revolutionized the way people and organizations interact with digital assistants. Large Language Models (LLMs), such as ChatGPT and DeepSeek, are versatile AI chatbots that utilize the Transformer architecture [6, 8]. They are developed using extensive datasets collected from the internet and have gained the capability to engage in conversations based on user questions [1, 2]. AI-driven conversational tools are increasingly employed across various domains, including academic research, business automation, customer service, and creative writing. These models undergo initial training on substantial volumes of text data and employ various training methods, including instruction regularization and reinforcement learning from human feedback (RLHF), among others.

In 2018, OpenAI launched the GPT-1 model, which was developed through unsupervised learning on a large corpus of text data. The subsequent release of GPT-2 in 2019 introduced notable enhancements in text generation, although it was initially withheld due to concerns about potential misuse. In 2020, GPT-3 debuted, featuring substantial improvements with its backing of 175 billion parameters, marking a significant advancement over earlier iterations. The release of GPT-3.5 in 2022 incorporated Reinforcement Learning from Human Feedback (RLHF). With the introduction of GPT-4, the model became more precise and dependable for a range of applications [4]. ChatGPT, developed by OpenAI, has become one of the most popular conversational AI models in the world. Its architecture supports various uses, such as customer service, technical assistance, education, and the automation of administrative duties. ChatGPT is well-known for its ability to engage in conversation, maintain coherence, and its extensive training on a variety of datasets. It has been engineered to produce text that mimics human language, respond to inquiries, and support various computational tasks [9, 10].

Furthermore, following the most recent launch of DeepSeek, which started in 2023 with DeepSeek V1, notable advancements were made in text comprehension and code generation, particularly in Chinese and English. Shortly after, DeepSeek-coder was introduced, optimized for software development, debugging, and addressing intricate programming challenges. In 2024, DeepSeek V2 was unveiled, featuring considerable enhancements in managing longer contextual interactions and improved logical reasoning [4]. DeepSeek AI aims to establish itself as a competitor by providing highly focused answers, particularly in specialized fields such as technical documentation, scientific studies, and specific problem-solving areas [10]. As these models gain traction, there is a growing need to evaluate their performance in real-world scenarios.

The objective of this study is to analyze how users perceive and evaluate the performance of ChatGPT and DeepSeek, two prominent advanced large language models (LLMs). It aims to provide a systematic comparison between ChatGPT and DeepSeek AI, highlighting aspects such as accuracy, usability, adaptability to various domains, and computational efficiency. By examining their respective strengths and weaknesses, this research aims to help users select the AI tool that best meets their needs. The results will enhance the existing body of literature on AI model comparisons and offer insights for future AI advancements.

For this evaluation, we employ deep learning (DL) models [11], such as Convolutional Neural Networks (CNNs), and Bidirectional Long-Term Memory (Bi-LSTM) to classify user-generated reviews in detail. These DL models help identify sentiment patterns, emotional subtleties, and common concerns expressed in textual feedback. Therefore, while LLMs are the focus of user comments, DL models serve as the methodological instruments for analyzing these discourses and comments. The main scientific contributions in this research are:

1. Our work introduced a novel comprehensive benchmark of ChatGPT and DeepSeek by combining subjective (LLM-based analysis) and objective (DL-based classification) approaches.
2. Developed a unified framework for sentiment and topic analysis utilizing TextBlob
3. Provided practical recommendations for enhancing AI app design and fostering user trust
4. Balanced Dataset Construction: Shows how RandomOverSampler can be used to correct the problem of class imbalance in the LLM user review datasets, so that training and evaluation can be made fair.
5. Deep Learning Benchmarking: Compares and benchmarks two popular deep learning models (CNN and Bi-LSTM) on sentiment classification on balanced LLM-related datasets, demonstrating that CNN is more accurate (96.41% vs. 93.12%).
6. LLM Improvement Insights: Provides functional and experiential gaps between DeepSeek and ChatGPT (e.g., connectivity and service problems vs. emotional connection and perceived intelligence), which can be used to improve future LLMs.
7. Methodological Framework: Suggests a unified framework that integrates qualitative LLM review analysis with quantitative deep learning classification, which can be replicated to benchmark emerging LLMs
8. Laid the groundwork for subsequent research on user experience with AI across various platforms

The rest of the paper is organized as follows: Section 2 explains why the given study is novel, whereas Section 3 presents the research questions. The related literature is reviewed in Section 4, and the methodology is discussed in Section 5. In Section 6, we develop a comparison of ChatGPT and DeepSeek using LLM-based sentiment analysis, and Section 7 provides an in-depth discussion of the insights developed. Section 8 presents the deep learning drive method of classification, and Section 9 provides the experimental results and discussion. Section 10 is a comparative review of the DL models on ChatGPT and DeepSeek reviews, and, finally, Section 11 concludes the study with the main findings and implications.

## 2 The Novelty in the Work

This paper proposes a dual-perspective system that combines the strength of sentiment analysis using an LLM (TextBlob, a lexical-based sentiment analysis tool) with deep learning-based binary classification (CNN and Bi-LSTM) models in the process of reviewing user opinions about ChatGPT and DeepSeek on the Google Play Store. In contrast to previous studies that have mainly employed either lexicon-based methods or deep learning analysis exclusively, this combination of both techniques offers a more reliable, balanced, and sophisticated approach to user perception and model performance. The combination of predictive classification and quantitative sentiment scoring helps address a missing element in the effective reception, functional aspects, and user experience dynamics of AI-driven applications by capturing both the emotional valence and structural patterns of user feedback.

## 3 Research Questions

This study aims to analyze and compare user perceptions of two leading LLM-powered applications—ChatGPT and DeepSeek—based on public reviews. The research focuses on how different sentiment analysis methods interpret user experiences with these apps. The core research questions are as follows:

**Table 1. Research Questions**

| List | Research Questions (RQ) |
|---|---|
| **RQ1** | What are the differences in user perceptions of ChatGPT and DeepSeek in terms of sentiment polarity (Positive, Neutral, Negative) based on real-world user reviews? |
| **RQ2** | Are deep learning models (CNN and Bi-LSTM) capable of classifying imbalanced user reviews after using balancing techniques like RandomOverSampler? |
| **RQ3** | Which deep learning model (CNN or Bi-LSTM) is more accurate, precise, recall, and F1-score in classifying sentiment across balanced datasets? |
| **RQ4** | What are the key linguistic and functional features that are mentioned in user reviews of ChatGPT and DeepSeek, and how do they help to understand user experience with Large Language Models (LLMs)? |

## 4 Literature Review

The widespread use of social media platforms has made SA an important research topic. As a result, a significant amount of writing about sentiment in this context has surfaced [12]. The introduction of ChatGPT in late 2022 marked the beginning of a new phase in Large Language Model (LLM) research. Since then, LLMs have progressed rapidly, with models such as GPT and Claude showcasing remarkable capabilities. Although open-source LLMs like LLaMA have obtained competitive outcomes based on specific metrics, their overall performance still falls short of that of proprietary models. In January 2025, DeepSeek made waves in the market and drew significant media attention with the launch of DeepSeek-V3 [13].

According to Wang et al. [14], Earlier research highlights the advantages and drawbacks of AI-based models, with a focus on their applications in areas such as customer service, content creation, and scholarly research. Recent studies have shown that while ChatGPT excels in general conversation and applications that require creativity, DeepSeek AI offers benefits in processing structured knowledge and solving problems in technical and scientific fields.

Previous studies by Bommasani et al. [15] highlight both the benefits and limitations of AI models, with a focus on their applications in sectors like customer service, content creation, and academic research.

According to Brown et al. [1], the rapid progress of Artificial Intelligence (AI) in Natural Language Processing (NLP) has led to the development of various AI-powered tools, including ChatGPT and DeepSeek AI. Most research on AI-driven conversational agents emphasises their language skills, flexibility, and efficiency in handling different tasks.

Wang et al. [14] proposed that an essential element of AI language models is their capability to comprehend and produce responses that resemble human communication. Research involving transformer-based models, such as GPT-4 and DeepSeek, demonstrates that both the model architecture and training datasets significantly influence performance.

Guo et al. [16] presented DeepSeek-R1a, a newly developed reasoning model that employs reinforcement learning to enhance contextual comprehension and decision-making capabilities. The complete model, which contains 671 billion parameters, has been supplemented with various distilled versions that have been made publicly available. Nevertheless, their efficacy in sentiment analysis has not yet been examined.

Trandabăț et al. [17] found that, according to public discussion, AI content lacks accountability, which raises essential concerns regarding who should be responsible when these AI systems produce misleading or dangerous outputs. Studies have investigated the public perception of Multiple Large Language Models (LLMs), yet researchers have not explored DeepSeek or its features and services within this scholarly context.

Others have examined users' opinions about more well-known LLMs, such as ChatGPT. In their study, Kabir et al. [18] looked at how people around the world feel about LLMs in languages that are not used much, focusing on how society as a whole feels about their spread. Koonchanok et al. [19]. Analysed Twitter data to determine user sentiment toward ChatGPT. They discovered that public reactions were primarily neutral to positive, with the primary concerns being cybersecurity, educational impact, and marketing applications.

According to Liao [20], who analyzed 67,000 Weibo posts, users showed both favorable opinions of ChatGPT in greater numbers but also raised some doubts regarding its impact on the education and employment fields.

## 5 Methodology

A comprehensive framework is presented for analyzing and evaluating users' emotional experiences using ChatGPT and DeepSeek, two large language models (LLMs) currently popular in the field. The process consists of four steps: data collection, data preparation, exploratory user data analysis, and sentiment classification using deep learning (DL) techniques. Figure 1 illustrates the methodology framework.

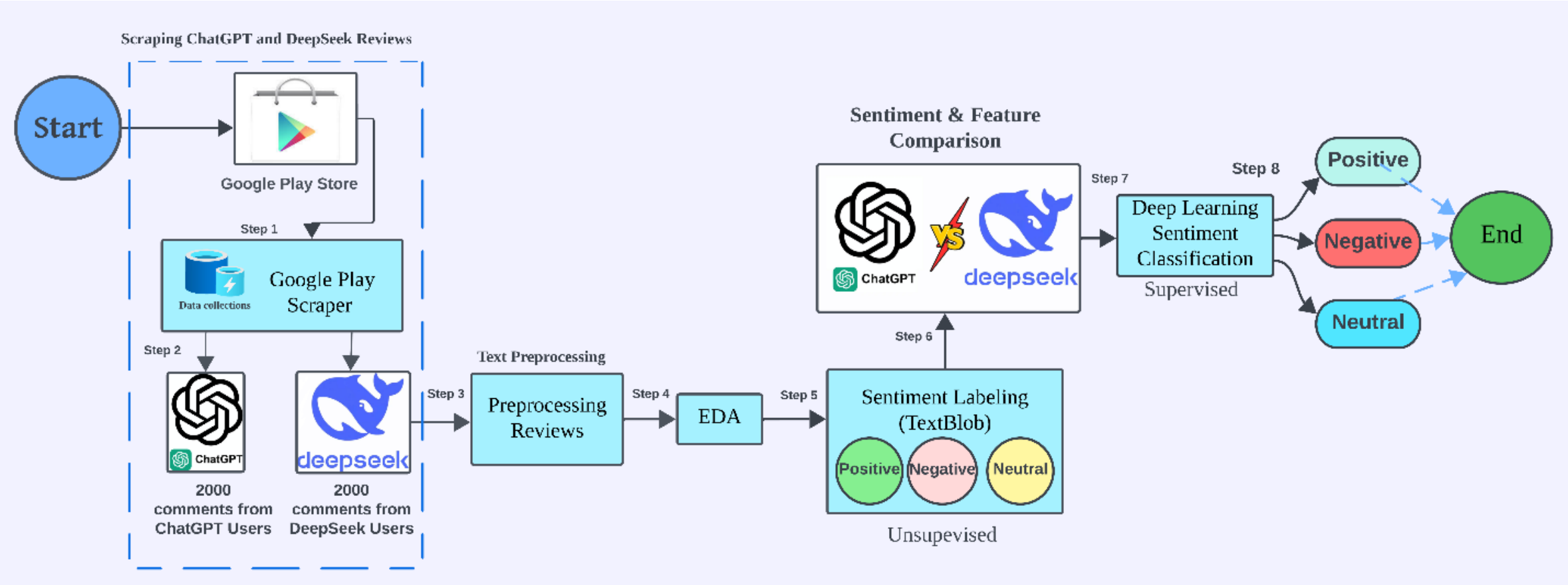

**Figure 1. Workflow of Sentiment Analysis and Classifying Comments from Users of ChatGPT and DeepSeek**

### 5.1 Data Collection, Processing, and Sentiment Classification for ChatGPT and DeepSeek

The initial data consisted of 4,000 user reviews gathered from ChatGPT and DeepSeek. The Distribution of classes was skewed, with some sentiment classes being underrepresented relative to others. Downsampling has been used in prior research to address the class imbalance in classes [21, 22, 23], but in this work, we used oversampling to retain all the instances in the majority class and produce a more balanced training set without the loss of information. This approach replicates samples of minority classes until each sentiment category has a similar number of samples. The oversampling of minority classes is slightly larger than it would be with the perfect doubling as a result of the synthetic sample generation and rounding [22]. As a result, the overall size of the dataset expanded to over 4,000 reviews. The final evaluation was conducted on 20% of the data (1,700 reviews), which served as the test set, representing 20% of the balanced data, not the original data, The test set consisted of 1,700 reviews, and the proportion of each of the sentiment classes (567 Negative, 567 Neutral, and 566 Positive reviews) was close to equal. The remaining 80% (~6,700 reviews) were used as training data, and 10% of the training data was held out as a validation set to monitor generalization during training (equations 1, 2, 3, and 4). This strategy ensures that the assessment is conducted on a balanced and representative sample, without bias towards the majority sentiment classes, and with reliable performance measures across the Negative, Neutral, and Positive classes.

1. Training Set (80%):

$$\mathcal{N}_{train_full} = 0.8\times\mathcal{N}_{total} \quad (1)$$

2. Test Set (20%):

$$\mathcal{N}_{test}=0.2\times\mathcal{N}_{total} \quad (2)$$

3. Validation Set (10%):

$$\mathcal{N}_{val}=0.1\times\mathcal{N}_{train_full}=0.1\times(0.8\times\mathcal{N}_{total}) = 0.08\times\mathcal{N}_{total} \quad (3)$$

4. Final Training Set (after removing validation set)

$$\mathcal{N}_{train}=\mathcal{N}_{train_full}-\mathcal{N}_{val}=0.8\times\mathcal{N}_{total}-0.08\times\mathcal{N}_{total}=0.72\times\mathcal{N}_{total} \quad (4)$$

5. Per Class Count:

$$\text{Per Class Count} = \frac{Test\ Set}{Number\ of\ classes} \quad (5)$$

## 5.2 ChatGPT Application

### 5.2.1 Scraping ChatGPT Reviews from the Google Play Store

To collect user reviews on ChatGPT, a massive scraping process was undertaken using the Google Play Scraper Python library. The targeted application was found through a unique application package name, openai.chatgpt.com, which the scraping is explicitly programmed to reflect the reviews written in English by users based in the United States. The reviews were then ordered by relevance to obtain the most informative ones. The data accumulation was carried out in iterative groups, with 100 reviews per request. Duplicating entries was removed to prevent redundancy and ensure high-quality data records. Duplicates were clustered by identical text content. This procedure was repeated until 2,000 unique reviews were obtained. The termination condition was automatically initiated when the scraper was no longer able to obtain more continuation tokens, indicating that all data about reviews had been exhausted. The latter was then organized as a Pandas DataFrame for further analysis.

### 5.2.2 Pre-processing and Lexicon–based Sentiment Labelling (TextBlob) for ChatGPT Reviews

After data acquisition, a standardized text pre-processing pipeline was applied to the collected reviews. The raw review texts were first converted to lowercase to ensure uniformity across all entries. Non-alphabetic characters were then removed to reduce noise and improve the performance of sentiment analysis models. The cleaned review texts were analyzed using the TextBlob library, which computes a polarity score ranging from -1 (entirely negative) to +1 (completely positive). Based on these polarity scores, sentiment labels were assigned according to the following thresholds: reviews with polarity scores above 0.1 were labelled as Positive, those below –0.1 as Negative, and those within the range of –0.1 to 0.1 as Neutral. The final labelled dataset was exported as a CSV file titled chatgpt_reviews.csv, providing a clean and structured resource for subsequent exploratory or predictive analyses. Figure 2 demonstrates the pre-processing steps of ChatGPT and DeepSeek.

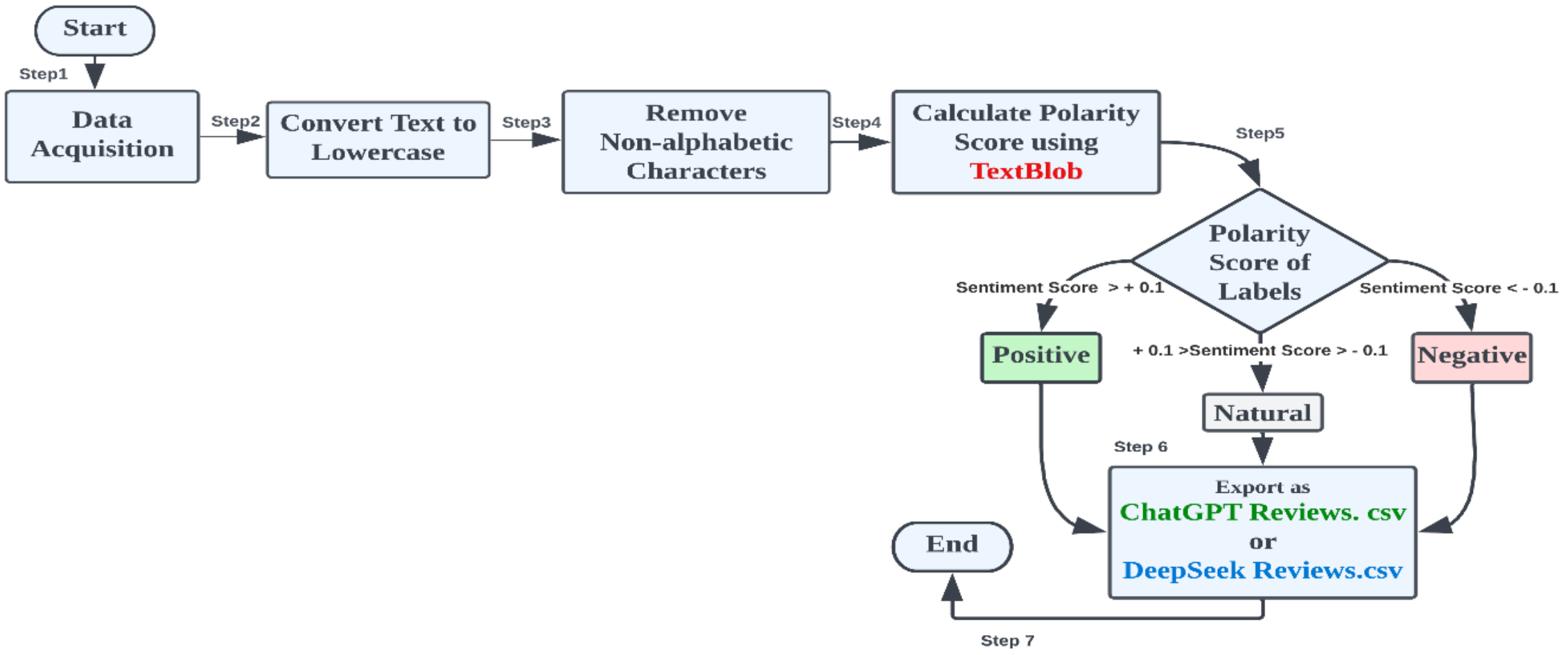

**Figure 2. Pre-processing and Lexicon–based Sentiment Labeling (TextBlob) for ChatGPT and DeepSeek Reviews**

### 5.2.3 Exploratory Data Analysis (EDA) for ChatGPT Reviews

In order to better analyse user impressions regarding ChatGPT, a data exploration (EDA) was performed to analyse the data gathered on the reviews [24]. This process served two main purposes.

First, a detailed analysis of the text-based user reviews was performed. Visualisations have been used to bring the data to life, depicting the Distribution of star ratings and sentiment classes. Word clouds were created for both positive and negative reviews to identify the most frequently used terms.

Second, EDA was statistically performed on the labeled dataset to examine the Distribution of the target class (“label” column) and to detect any class imbalance. This was a crucial step to ensure that any imbalance, which could negatively impact model performance, was addressed early in the process. RandomOverSampler techniques were employed to balance the dataset. We applied fundamental data cleaning steps, including the elimination of duplicate entries, to ensure data integrity.

### a) Distribution of Star Ratings

In this section, there is a presentation on the Distribution of star ratings provided by individuals who use the ChatGPT app. The visualization may indicate the level of total user satisfaction and provide an idea about the proportion of positive, neutral, and negative feedback.

Figure 3 displays the given star-rating Distribution. The distinct high at the 5-star mark represents overall high satisfaction among users, while the lowest represents the opposite. Although there are a few low ratings, the picture of the Distribution shows the skewness of the Distribution to the right, indicating that the app was rated positively by many users.

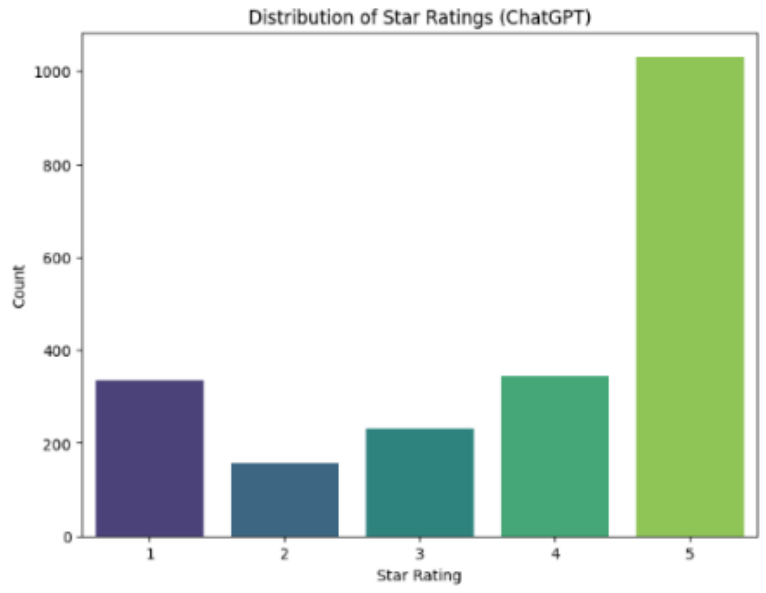

**Figure 3. Distribution of Star Ratings for ChatGPT Reviews**

### b) Sentiment Distribution

Based on the sentiment analysis (see Figure 4), the most popular reviewing sentiment was positive (0.6935 reviews), and the quantity of entries marked as neutral (0.2340 reviews) and as negative (0.0724 reviews) constituted a relatively small proportion. This tendency has been associated with a different distribution of ratings, thereby establishing a positive perception of the ChatGPT application among most users.

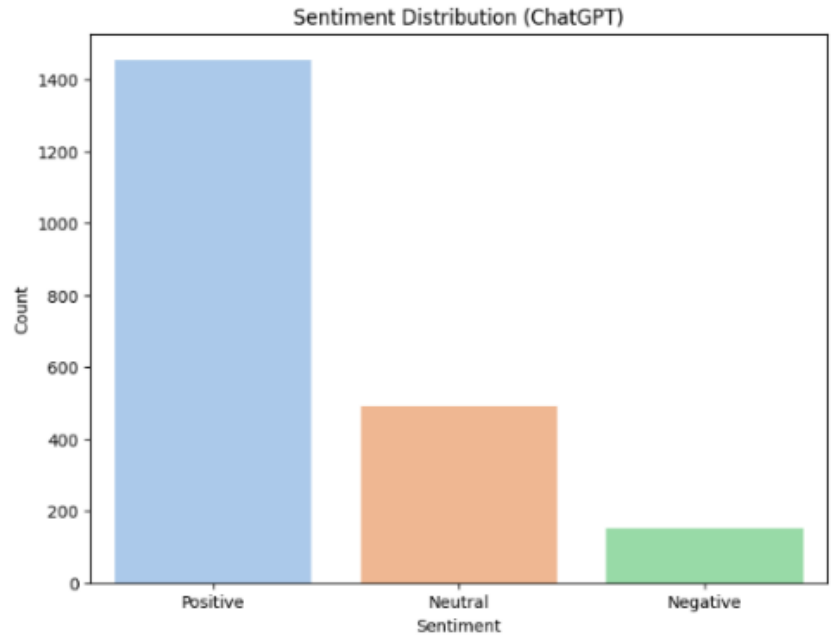

**Figure 4. Sentiments Distributions**

### c) Word Cloud of Positive Reviews

The positive reviews dataset has been used to make a word cloud, which is presented in this subsection. The use of dominant terms such as "*good*", "*help*", "*feature*", "*answer*", "*use*", "*helpful*", "*response*", "*useful*", "*easy*", *and* "*give*" defines the characteristics that satisfied users like the most. Figure 5 illustrates a Word Cloud of Positive Reviews for the ChatGPT application.

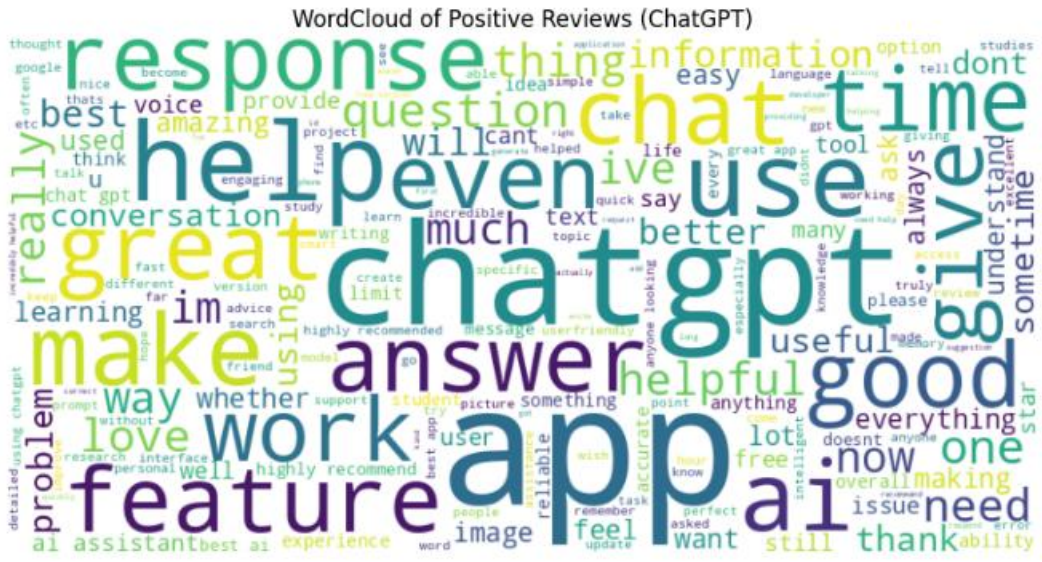

**Figure 5. Word Cloud of Positive Reviews for ChatGPT**

### d) Word Cloud of Negative Reviews

The subsection contains a word cloud representation of the collection of negative user reviews (corpus). The most common of those terms are *“app”, “wrong”, “time”, “can't", "issue", and "message"*, which show the principal concerns and pain points of the users. Figure 6 illustrates a Word Cloud of Negative Reviews for the ChatGPT application.

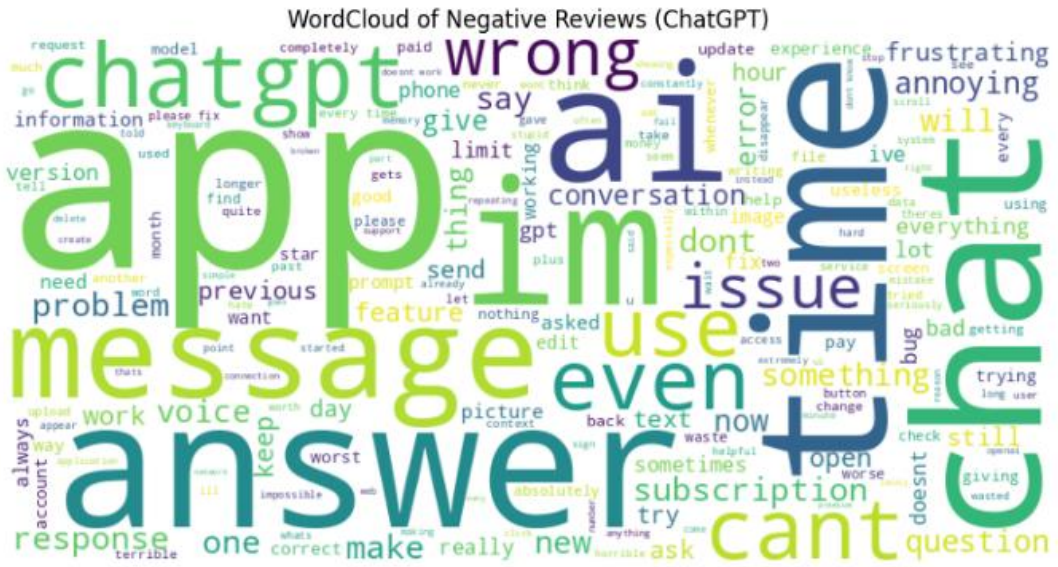

**Figure 6. Word Cloud of Negative Reviews for ChatGPT**

## 5.3 DeepSeek Application

### 5.3.1 Scraping DeepSeek Reviews from Google Play Store

To assess public sentiment toward the DeepSeek application, user reviews were systematically collected from the Google Play Store using the Google Play Scraper library in Python. The application was uniquely identified by its package name, deepseek.chat.com, and the data extraction was configured to target English-language reviews submitted by users located in the United States. The scraping procedure employed a batch-wise retrieval approach, fetching reviews in groups of 100 entries per request. These reviews were sorted by relevance to ensure that the most meaningful and representative feedback was prioritised. Each new batch was appended to a cumulative list of reviews until a total of 2,000 entries was collected. To ensure completeness, the process automatically terminated upon depletion of continuation tokens, signaling that all accessible reviews had been retrieved. The resulting corpus was then formatted into a structured DataFrame using the Pandas library to facilitate further processing.

### 5.3.2 Pre-processing and Lexicon–based Sentiment Labelling (TextBlob) for DeepSeek Reviews

After extracting the data through acquisition processes, a common pre-processing pipeline was applied to the recurring text of the reviews to increase consistency and prepare the content for analysis. This pre-processing involved the elimination of punctuation as well as numeric figures and any other non-alphabetic characters. Also, the capitalisation of all the text was changed to lowercase in order to unify the format. The cleaned text was then analysed in terms of sentiment using the library TextBlob, which provides the polarity score in the range of 1.0 (most positive) to -1.0 (most negative). Reviews were classified into the three sentiment classes, namely, positive (polarity > 0.1), negative (polarity < -0.1), and neutral (0.1). The end-labeled dataset in the form of precisely 2000 reviews was saved in the CSV file named deepseek_reviews.csv. The given dataset served as the source for further comparative analysis, exploration, and model training. See Figure 2.

### 5.3.3 Exploratory Data Analysis (EDA) for DeepSeek Reviews

An Exploratory Data Analysis (EDA) was conducted to uncover key patterns in user feedback on the DeepSeek application. This phase focused on analysing the Distribution of star ratings and Cloud Word, which present the types of users' sentiments, whether Positive or Negative, about the DeepSeek application.

### a) Distribution of Star Ratings

In Figure 7, the star ratings given by the users are outlined. Like ChatGPT, the ratings given by the majority of users were 5-star, which amounts to high satisfaction. However, there is presence of this is indicated by a conspicuous number of 1-star ratings indicating a level of polarisation in user experience.

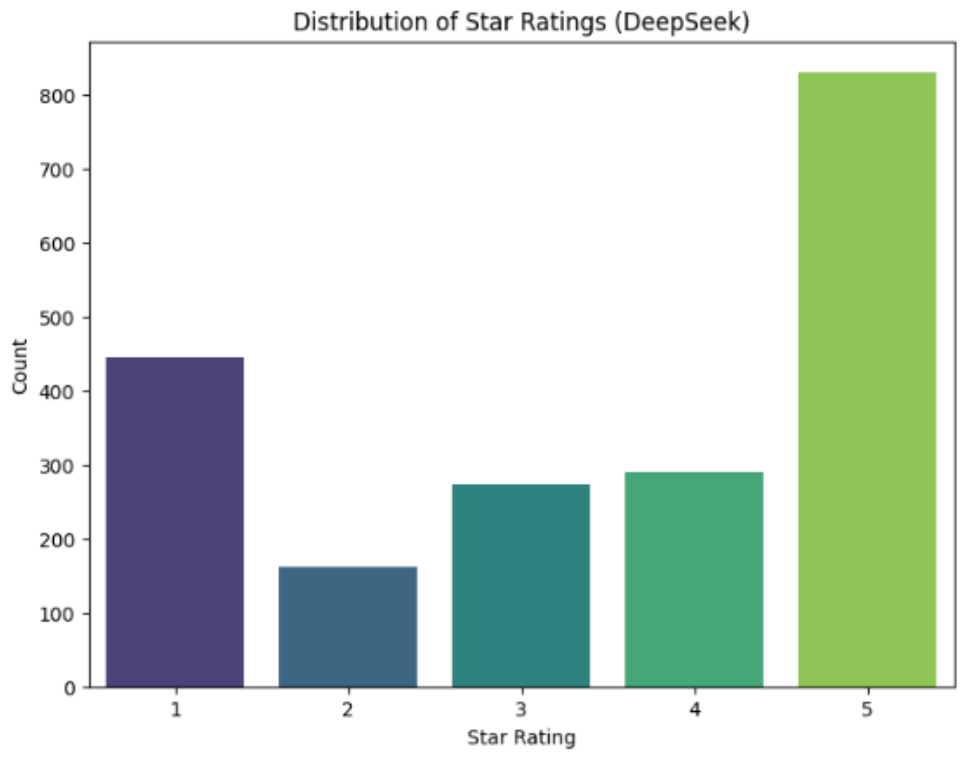

**Figure 7. Distribution of Star Ratings for DeepSeek Reviews**

### b) Sentiments Distibutions

The dataset reflected in Figure 8 can be analyzed, and it shows that positive reviews (0.6885 reviews) are the most frequent, followed by neutral reviews (0.2315 reviews), and negative reviews (0.0800 reviews) are the least frequent group, in terms of their frequency. Such a sentiment distribution is consistent with the general Distribution of stars, which suggests even more reasons to believe that user ratings are mostly positive.

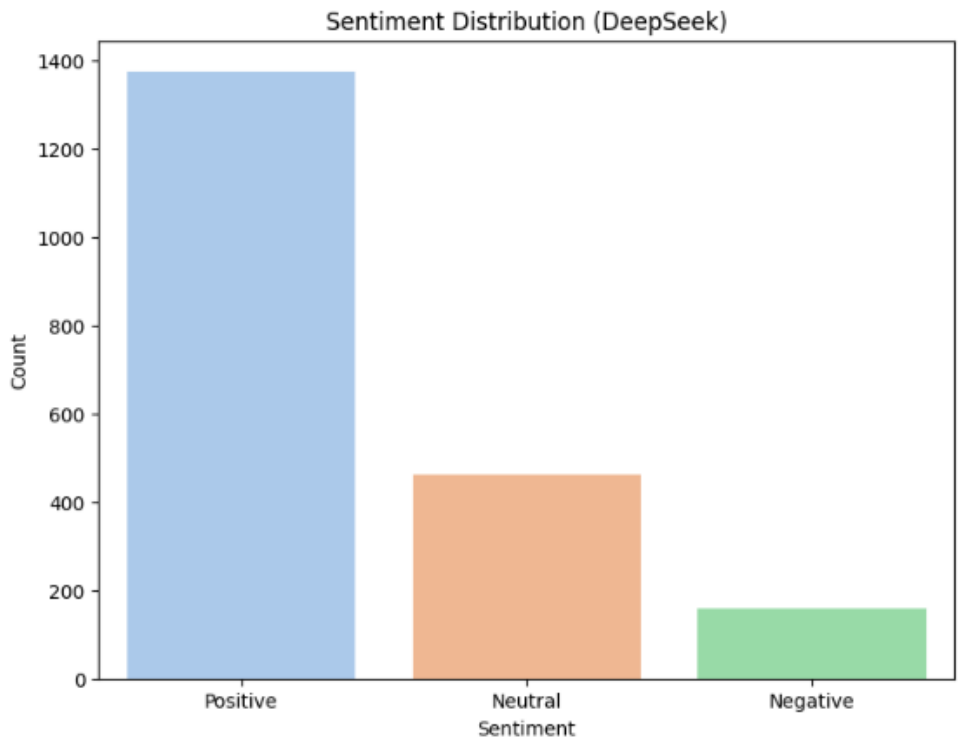

**Figure 8. Sentiment Distribution in DeepSeek Reviews**

### c) Word Cloud of Positive Reviews

The analysis of a positive user feedback set regarding this app has been visualised in a word cloud, where common phrases and words like *"answer", "DeepSeek", "free", "good", "better", and "amazing"* have determined the features delivered by this application to the satisfaction of the users. Figure 9 shows the positive reviews for the DeepSeek app.

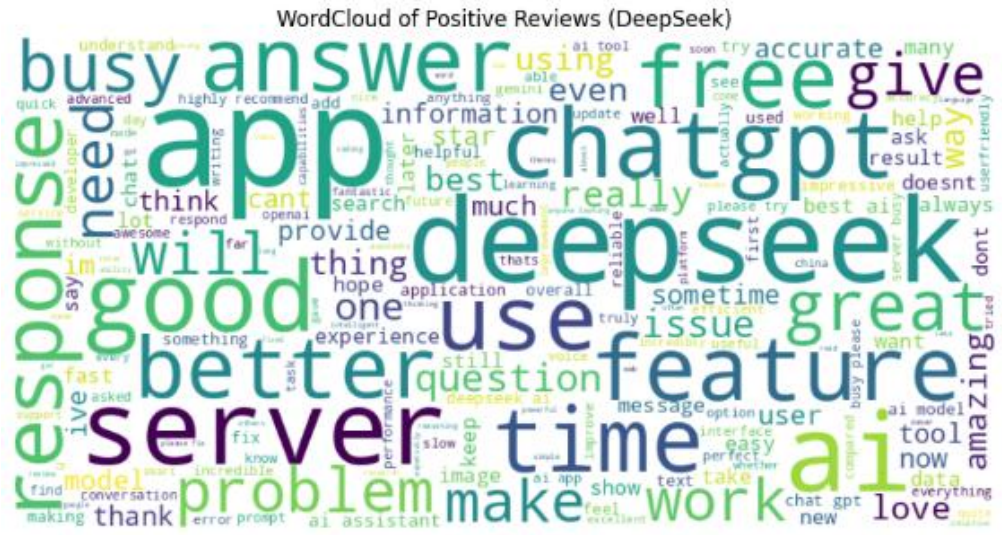

**Figure 9. Word Cloud of Positive Reviews for DeepSeek**

### d) Word Cloud of Negative Reviews

The subsection contains a cloud word with the total negative rating from users, one of the concerns and vulnerable aspects of this application, as noted by users. The commonly used words and phrases that we have seen were "time", "issue", "bad", "can't", "busy", "frustrating", "don't", "unless", "error", and "*slow"*. Figure 10 demonstrates the negative reviews for the DeepSeek app.

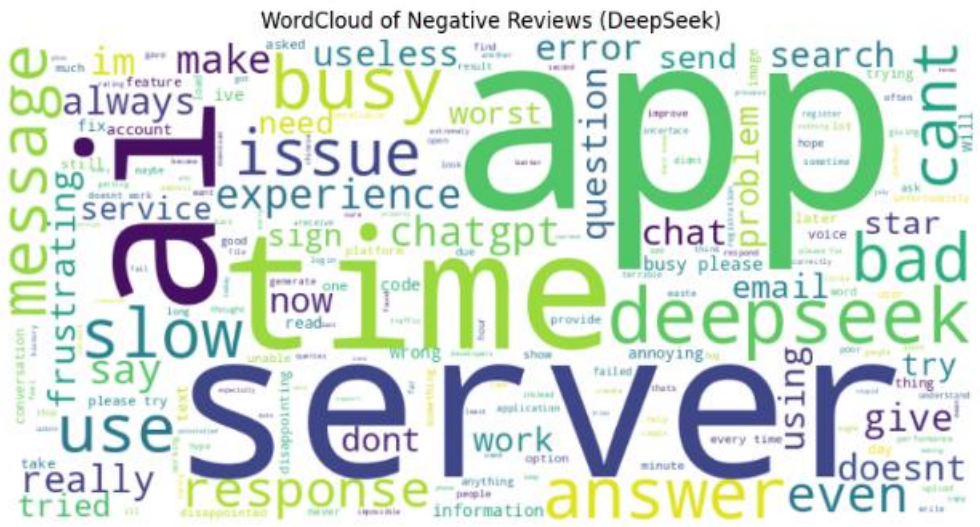

Figure 10. Word Cloud of Negative Reviews for DeepSeek

## 6 ChatGPT vs. DeepSeek

This section presents a comparative profile of ChatGPT and DeepSeek, focusing on user generation. Retrieve statistics compiled at the Google Play Store. The review is conducted in terms of rating stars and sentiment distribution. Distribution of stars in both applications.

In Figure 11 and Tables 2 and 3, the Distribution of sentiments in a sentiment analysis reveals that both ChatGPT and DeepSeek exhibit predominantly positive feelings, with 69.35% and 68.85% of their responses, respectively. Neutral mood accounts for approximately 23% in both models, with negative outputs being minimal at 7.25% and 8.00%, respectively, in ChatGPT and DeepSeek. The patterns in this regard indicate a similar inclination towards positivity, although ChatGPT has a slightly lower negative sentiment rate, most likely due to model tuning or training data.

**Table 2. Comparison between ChatGPT and DeepSeek for Sentiment Analysis of Users**

| LLMs Models | Sentiments Distributions | | |
|---|---|---|---|
| | Positive | Natural | Negative |
| **ChatGPT** | 0.693518 | 0.234032 | 0.072450 |
| **DeepSeek** | 0.688500 | 0.231500 | 0.080000 |

**Table 3. Comparison between ChatGPT and DeepSeek for Star Rating**

| LLMs Models | Average Score of Star Rating |
|---|---|
| **ChatGPT** | 3.754051 |
| **DeepSeek** | 3.450500 |

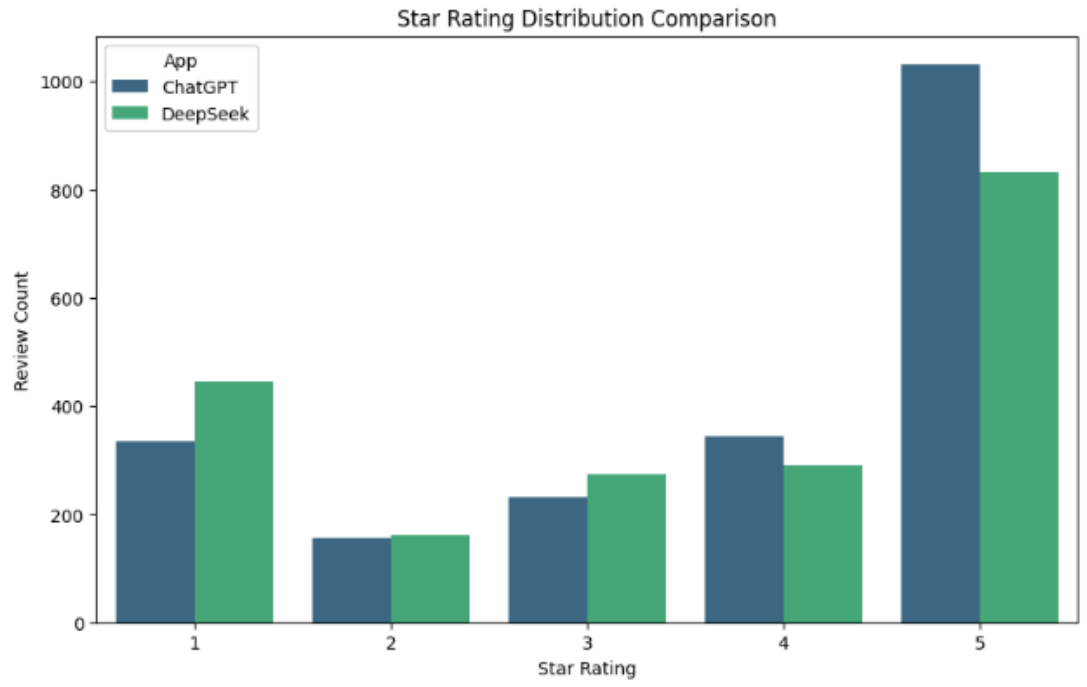

Figure 11. Star Rating Distribution Comparison for proposed LLM models

Figure 12 also shows that ChatGPT had an average star rating of 3.75, compared to DeepSeek, which had an average of 3.45 stars. This represents a relatively small gap but suggests higher satisfaction among ChatGPT users. This fact indicates that ChatGPT is more likely to meet user expectations regarding the quality of interaction and perceived usefulness.

The sentiment distribution analysis shown in Figure 13 indicates that both LLMs produce nearly identical percentages of positive responses; ChatGPT yields 69.4%, and DeepSeek yields 68.9%. The rate of neutral output is 23.4% for ChatGPT and 23.2% for DeepSeek, whereas negative sentiments are relatively small, at 7.2% and 8.0% for ChatGPT and DeepSeek, respectively. Summarizing the findings, it can be inferred that there is a general disposition towards positivity in both LLMs, with ChatGPT exhibiting a slight superiority by being more positive and less harmful, and rating higher among users. These differences may be explained by divergent strategies related to fine-tuning, other mechanisms of safety alignment differences, or changes in training data construction.

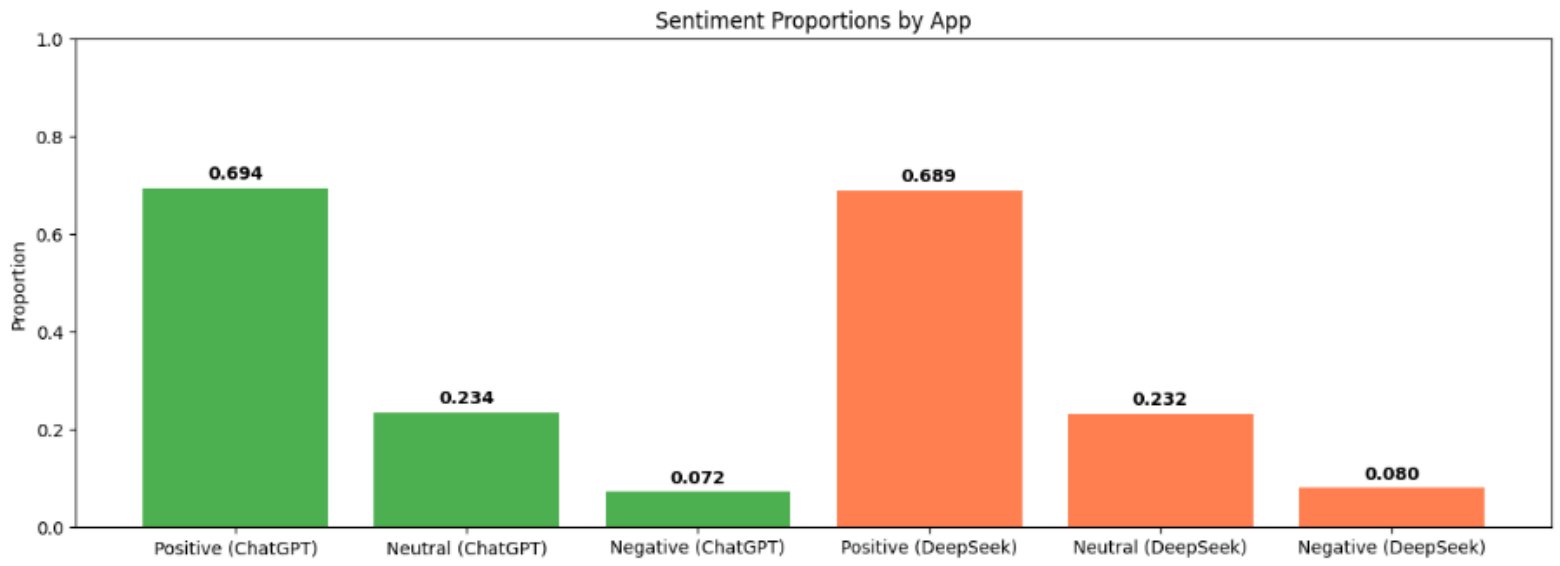

**Figure 12. Sentiment Proportions by App**

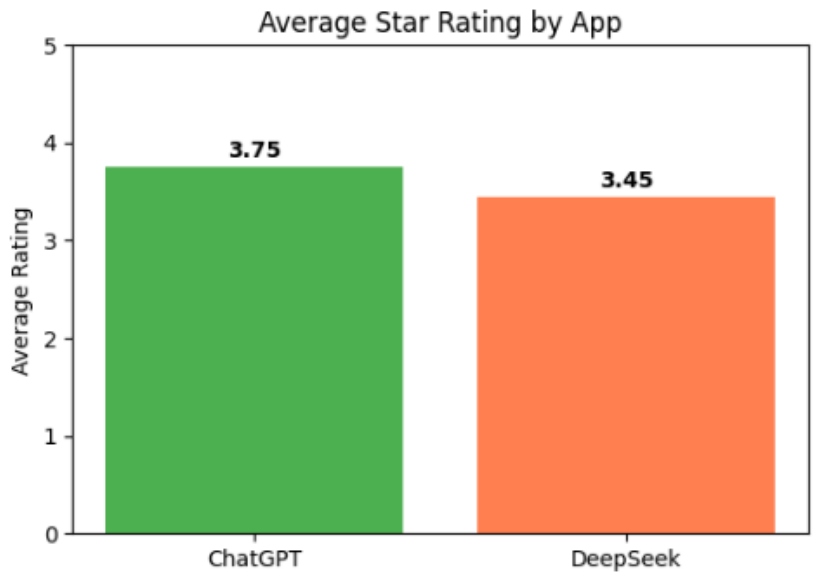

**Figure 13. Star Rating Distribution Comparison for proposed LLM models**

The comparison, based on a corpus analysis of the 20 most common lexical items that reflect frequency of occurrence, summarised in Figure 14, shows that lexical divergence is pronounced. ChatGPT shows higher counts for terms like "app" and "ChatGPT", indicating greater emphasis on application-related discussions and brand mentions. DeepSeek stands out in having a high rate of the cues "DeepSeek", "even", and "busy", suggesting a stronger focus on operational or situational contexts. The lexical overlap can be observed in words like 'time' and 'use', as they demonstrate the regularities of discourse in the two corpora. Such patterns are most likely to emerge as a result of contrasted training data and model parameters.

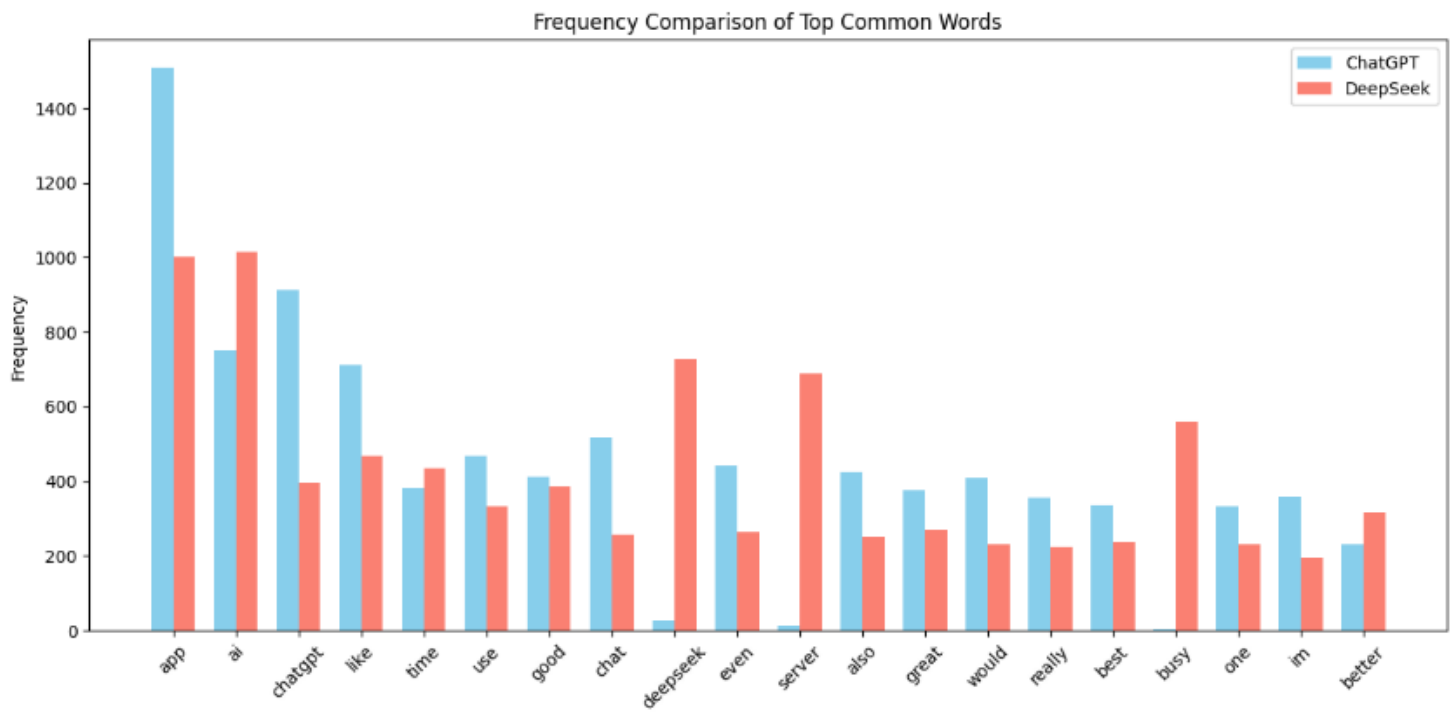

**Figure 14. Frequency Comparison of Common Words Between ChatGPT and DeepSeek**

## 7 Analysis and Insights

An evaluation comparison shows that ChatGPT currently rates higher and more positively than DeepSeek. Feedback to ChatGPT refers to the notion of perceived usefulness, smartness, and general utility. Assessments of DeepSeek have a larger ratio of negative remarks, the most common being caused by the inability to connect to the server, as well as technical complications. Language analysis reveals that the DeepSeek review primarily focuses on functional and service-related topics, whereas the ChatGPT report incorporates elements of service quality and personal user experience, resulting in more extensive, emotional, and generally positive text. Collectively, these results suggest potential areas for infrastructural improvement in DeepSeek and its continuation, focusing on greater user-centric sophistication.

## 8 Classifications Using DL Models

Finally, the models of deep learning were developed and trained to distinguish between the sentiment of user reviews. The number of words in the text data was tokenised and padded so that each call could have the same length. Sentiment labels were assigned to numeric values, and the dataset was balanced using the RandomOverSampler method. To treat the issue of class imbalance. Two neural network structures were implemented: a Bidirectional Long Short-Term Memory (Bi-LSTM) model, followed by global max pooling, and a Convolutional Neural Network (CNN) model. Every model utilised an embedding layer, a task layer, and a final SoftMax output, a three-class model.

### 8.1 Convolutional Neural Networks (CNN)

The suggested CNN prescribed by a network consists of an embedding layer, which transforms input token strings into dense vectors. In this model, words are represented as vectors in a 64-dimensional space that maps the vocabulary of up to 5,000 words; this design ensures that semantically related words are mapped into similar vectors. Then, there is a one-dimensional convolutional layer with 128 filters, using a kernel size of 5 to move along the sequence and extract local n-gram patterns. The application of the ReLU activation layer enables the network to learn complex text features by introducing non-linearity into the model.

Then, the Global Max Pooling layer [25] is used to identify the most descriptive feature in each feature m, thereby carrying out dimensionality reduction while preserving vital information. The batch output is then fed to a fully connected dense additional layer of 64 units with ReLU activation to build more abstract feature representations. In order to reduce the chance of overfitting, a dropout layer with a probability of 0.5 is added, which randomly disables a fraction of the neurons during training.

Lastly, the model makes predictions via a softmax layer with three neurons, representing the sentiment classes: Negative, Neutral, and Positive. Essentially, this architecture demonstrates a substantial ability in extracting functional textual characteristics and classifying them with great success into sentiment categories. See Figure 15. The steps below provide a general account of how the model is structured and how it works (Algorithm 1):

**Algorithm 1:** Pseudocode of CNN Model for Text Classification

1: **Start**
2: **Initialise** the model:
  model ← Sequential ()
3: **Add** embedding layer:
  model. add (Embedding (input_dim = max_words, output_dim = embedding_dim, input_length = max_length))
4: **Add** a convolutional layer:
  model. add (Conv1D (filters = 128, kernel_size = 5, activation = 'relu'))
5: Apply global max pooling:
  model. add (GlobalMaxPooling1D ( ))
6: **Add** fully connected layers:
  model. add (Dense (64, activation = 'relu'))
  model. add (Dropout (0.5))
  model. add (Dense (3, activation = 'softmax'))
7: **Compile** the model:
  model. compile (loss = 'sparse_categorical_crossentropy', optimizer = 'adam', metrics = ['accuracy'])
8: **Train** the model with early stopping:
  model.fit (X_train, y_train, epochs = 50, batch_size = 32, validation_split = 0.1, callbacks = [early_stop])
9: **Evaluate** the model:
  model. evaluate (X_test, y_test)
10: **End**

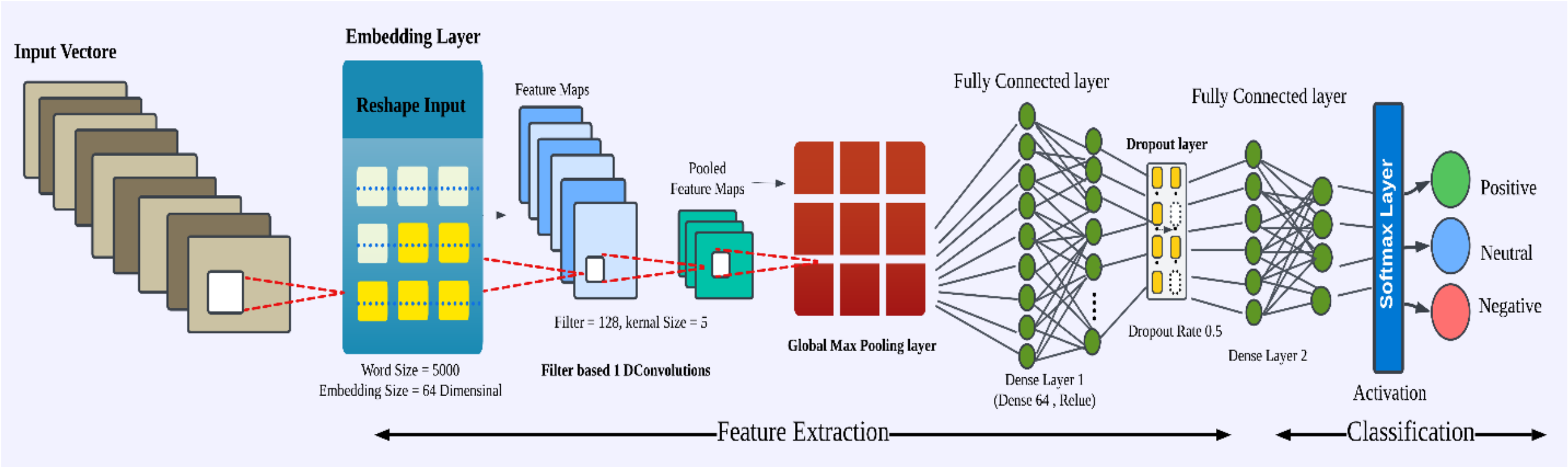

**Figure 15. CNN Architecture**

## 8.2 Bidirectional Long Short-Term Memory (Bi-LSTM)

The similarities between Bi-LSTM and the rest of the recurrent neural networks are that they enable interactions with the model and allow users to extract valuable information from the orderings of text sequences that move back and forth in time. The model has the first stage, the buying formula, which consists of buying motives and environmental antecedent conditions. The conversion of the characters to a particular set of sequences was done by transforming them into vectors of a 5,000-token vocabulary. Combined with the embedding parameter of 64, it allows the semantic encoding process in this layer. The statistics are processed through a bi-directional LSTM layer of 64 memory units [24]. In its two orientations, Standard LSTM layers contrast with their bidirectional version due to the different purpose of the background color in these two versions. This variant takes a different sequence processing path by examining in forward and backward word directions simultaneously. Due to the nature of the implemented task, Bi-LSTM performs exceptionally well in sentiment analysis. The majority of the work also spans both forward and backwards directions in the treatment of complete sentence contexts. The Bi-LSTM layer has a 50% dropout layer as input to prevent overfitting.

The model uses ReLU activation in the dense layer of 64 neurons, where it transfers the information. Additionally, it can identify non-linear relationships when working with complex extracted features. To enhance robustness in the model, an additional layer of dropouts is included. The 3-unit dense output layer takes the softmax function. The percentage of Negative, Neutral, and Positive sentiment classifications was thus generated in terms of probabilities, arrived at by the activation of this model. The design of the architecture incorporates linear forms of understanding, featuring extraction—the methods used to obtain more efficient results in complicated language processing tasks. The following gives a general overview of the model structure and functioning process: (Algorithm 2, and see Figure 16).

---

**Algorithm 2:** Pseudocode of Bi-LSTM Model for Text Classification

---

**1: Start**
2: **Initialise** the model:
  model ← Sequential ()
3: **Add** embedding layer:
  model.add (Embedding (input_dim = max_words, output_dim = embedding_dim, input_length = max_length))
4: **Add** recurrent layer:
  model. add (Bidirectional (LSTM (64)))
5: **Apply** dropout regularisation:
  model. add (Dropout (0.5))
6: **Add** fully connected layers:
  model. add (Dense (64, activation = 'relu'))
  model. add (Dropout (0.5))
  model. add (Dense (3, activation = 'softmax'))
7: **Compile** the model:
  model. compile (loss = 'sparse_categorical_crossentropy', optimizer = 'adam', metrics = ['accuracy'])
8: **Train** the model with early stopping:
  model.fit (X_train, y_train, epochs = 50, batch_size = 32, validation_split = 0.1, callbacks = [early_stop])
9: **Evaluate** the model:
  model. evaluate (X_test, y_test)
10: **End**

---

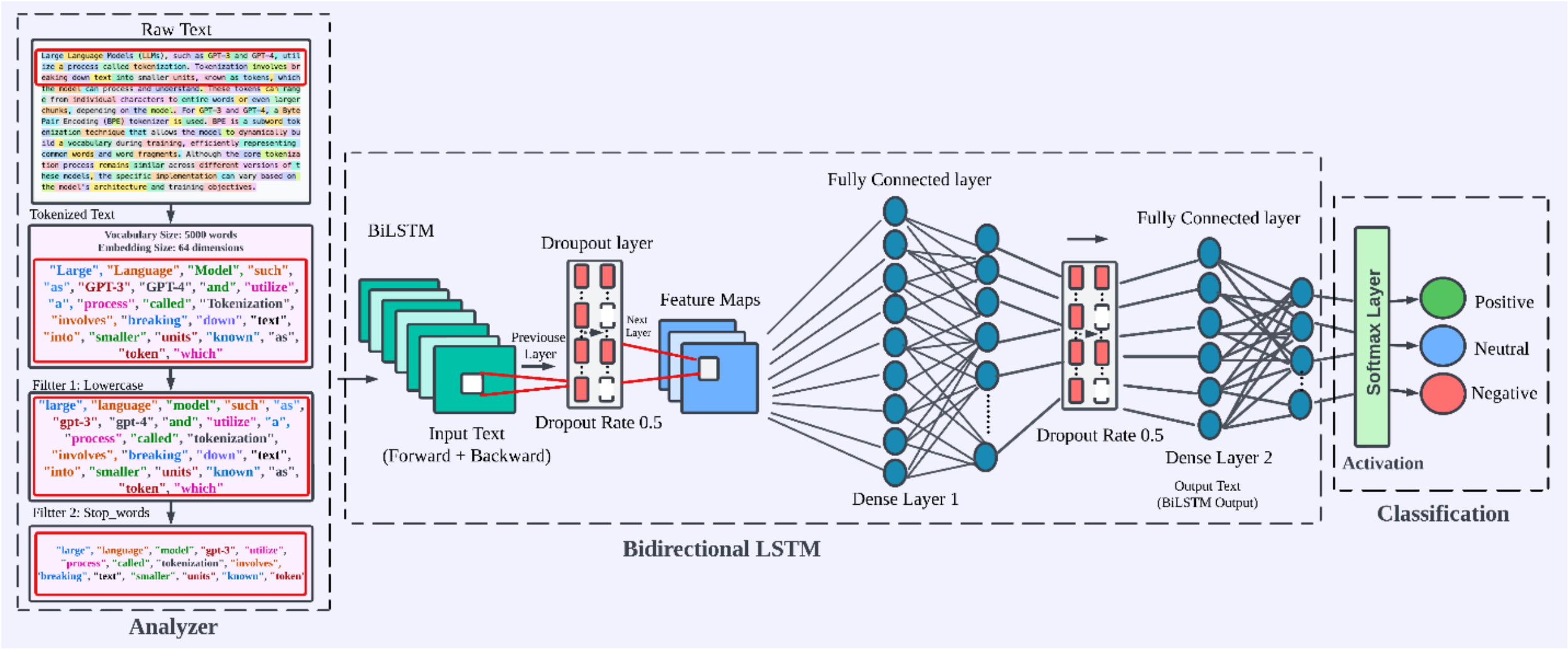

**Figure 16. Bi-LSTM Architecture**

## 9 Results and Discussions

Evaluating the performance of the proposed deep learning model using a balanced set of 1,700 test samples that were equally divided into Positive, Neutral, and Negative classes. It is noteworthy that, although the descriptive analysis compared the ChatGPT and DeepSeek reviews in terms of polarity and user complaints, the deep learning classifiers were trained on the combined dataset to assess overall sentiment classification performance. These models were not aimed at comparing applications, since the effectiveness of the deep learning approach to sentiment classification in the context of applications with LLM is confirmed to be Positive, Neutral, and Negative. The overall accuracy, loss, and classification metrics are summarised in Tables 4 and 5.

**Table 4. Performance of Proposed DL Models**

| DL Models | Accuracy% | Test Loss |
|---|---|---|
| CNN | 96.41 | 0.1178 |
| Bi-LSTM | 93.12 | 0.2522 |

**Table 5. Classification Report of Proposed Models**

| DL Models | Precision% | | | Recall% | | | F1-Score% | | |
|---|---|---|---|---|---|---|---|---|---|
| | Positive | Neutral | Negative | Positive | Neutral | Negative | Positive | Neutral | Negative |
| CNN | 0.99 | 0.95 | 0.95 | 1.00 | 0.94 | 0.95 | 0.99 | 0.95 | 0.95 |
| Bi-LSTM | 0.99 | 0.91 | 0.89 | 0.99 | 0.88 | 0.92 | 0.99 | 0.90 | 0.91 |

### 9.1 Conventional Neural Network (CNN)

Tables 4 and 5, as well as Figures 17 and 18, indicate that the CNN model performed best on the balanced dataset, achieving a test accuracy of 96.41% and a loss of 0.1178. The model demonstrated high predictive ability across all categories of sentiment.

In the Negative class, CNN achieved a Precision of 0.95, a recall of 0.95, and an F1 of 0.95, which is a strong result in terms of identifying negative reviews. In the Neutral category, CNN achieved a Precision of 0.95, a recall of 0.94, and an F1 score of 0.95, indicating a high capacity to identify neutrality despite the linguistic overlap with other sentiments. The Positive class produced the best results with a Precision of 0.99, a Recall of 1.00, and an F1-score of 0.99, indicating that CNN was nearly perfect in detecting positive sentiment. The confusion matrix also suggests the model's effectiveness: a few Neutral reviews were incorrectly classified as Negative (6 samples), and 28 Positive reviews were incorrectly classified as Neutral. This shows that CNN generalizes well and can leverage its capability to capture local n-gram features through convolutional filters, which helps reduce sentiment ambiguity.

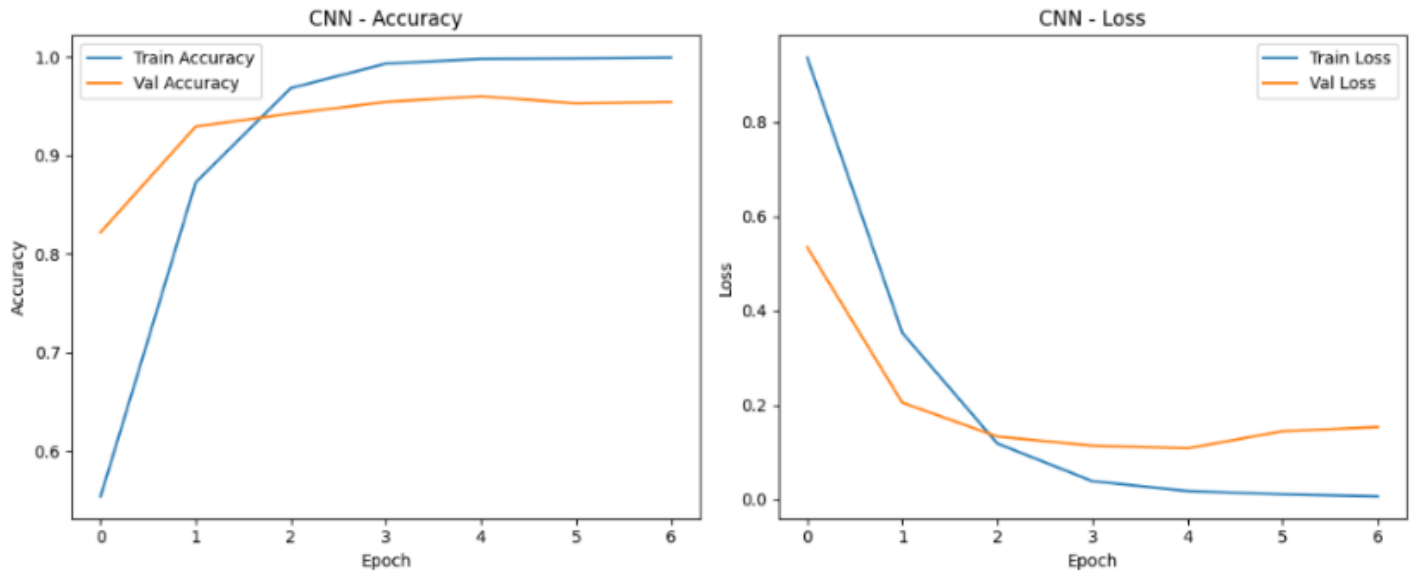

**Figure 17. Accuracy of CNN Models**

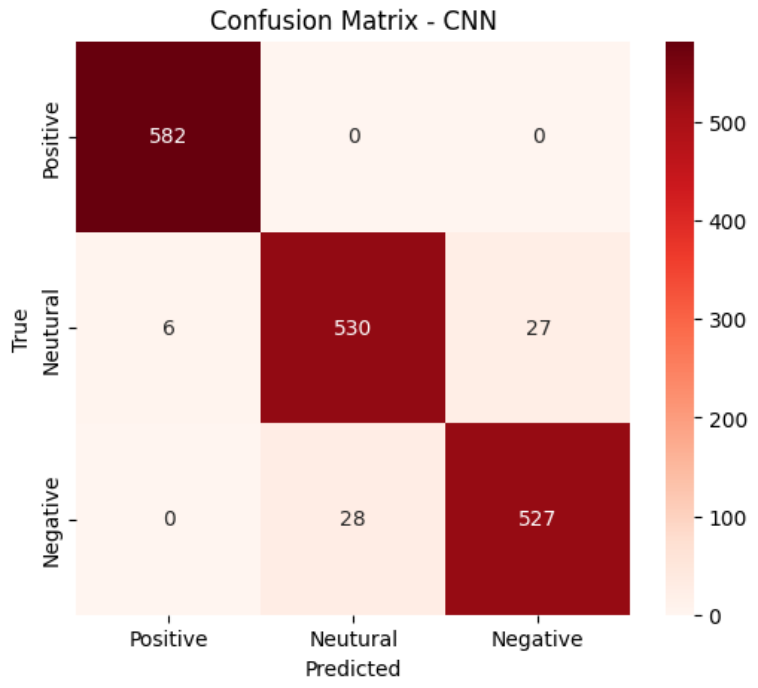

**Figure 18. The Confusion Matrix of CNN Models**

## 9.2 Bidirectional Long-Short Term-Memory (Bi-LSTM)

On the same balanced dataset of 1,700 reviews, the Bidirectional LSTM model had a test accuracy of 93.12% and a loss of 0.2522. Tables 4 and 5, as well as Figures 19 and 20, indicate that the performance was competitive, but slightly lower than that of CNN. The Negative class performed well (Precision of 0.89, recall of 0.92, F-score of 0.91), but not as well as CNN. The Neutral class had a Precision of 0.91, a recall of 0.88, and an F1 Score of 0.90, indicating that it was moderately difficult to distinguish between Neutral and Positive sentiment. The Positive category was also powerful, with a Precision of 0.99, a Recall of 0.99, and an F1 score of 0.99, indicating almost perfect classification in this category. The confusion matrix showed that the misclassification rate was higher than that of CNN, especially between Neutral and Positive reviews. In particular, 60 Positive samples were incorrectly classified as Neutral, and 44 Neutral samples were classified as Positive, which suggests that Bi-LSTM was unable to distinguish between these two classes clearly due to overlapping linguistic patterns.

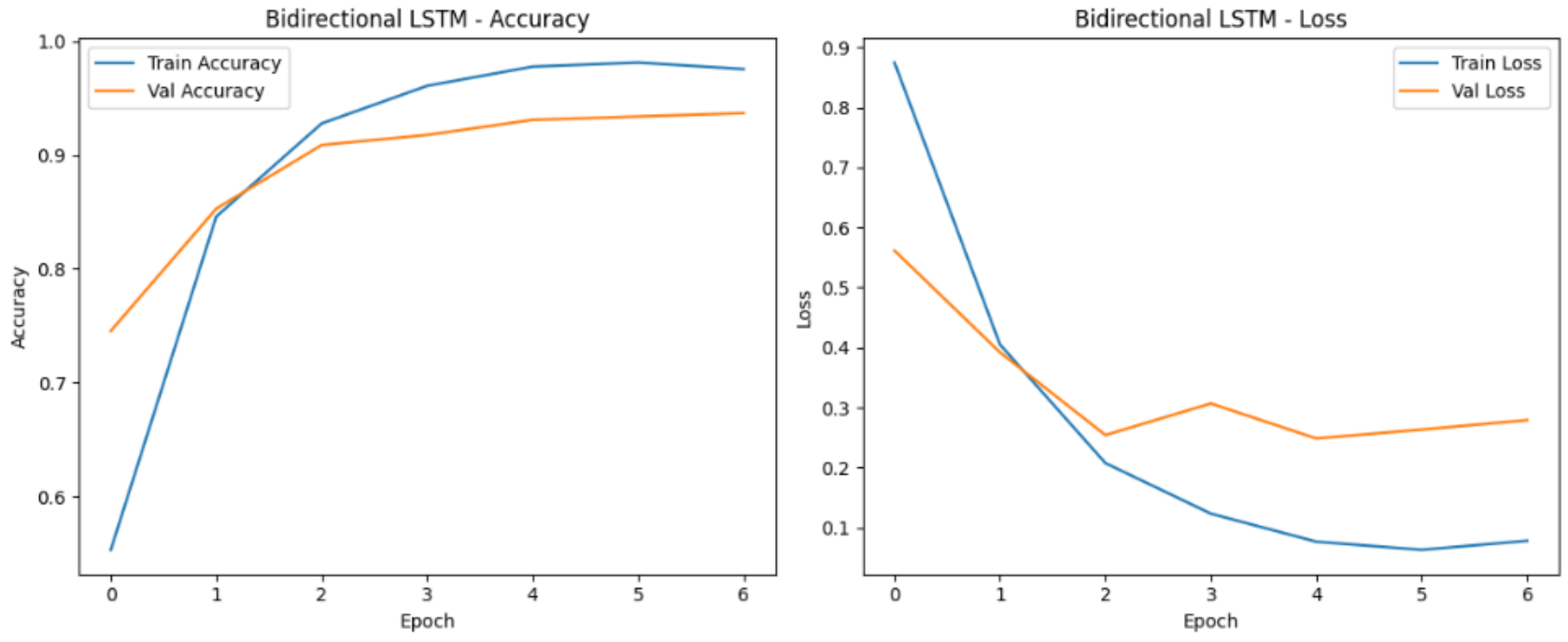

**Figure 19. Accuracy of Bi-LSTM Models**

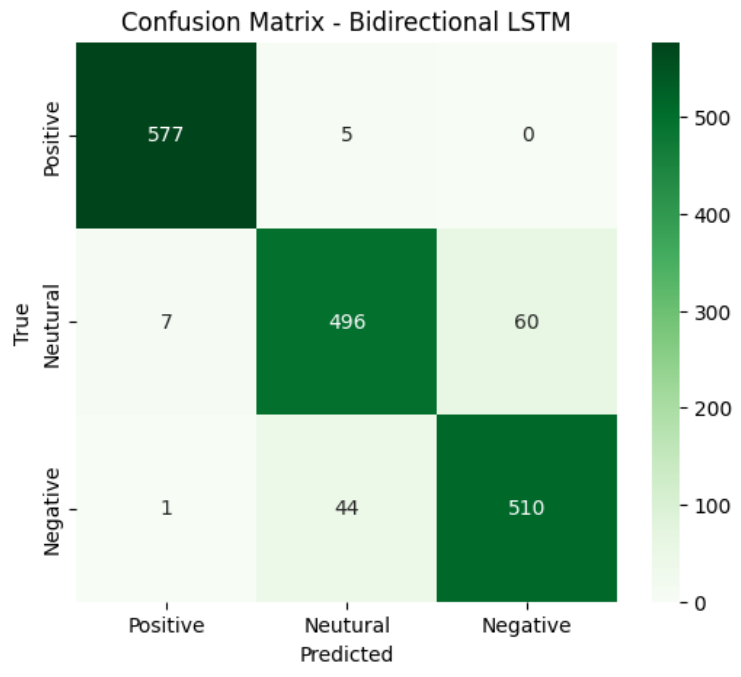

**Figure 20. Accuracy of Bi-LSTM Models**

In answering the research questions, the study reveals evident differences in the perception of users of large language models (LLMs), such as ChatGPT and DeepSeek. ChatGPT was primarily reviewed positively, with a focus on its usefulness, intelligence, and overall quality of interaction. In contrast, DeepSeek was reviewed more negatively, with a focus on server connectivity issues and technical limitations. In terms of deep learning classification performance, the comparative analysis showed that CNN performed better (96.41% accuracy) than Bi-LSTM (93.12%), especially in classifying Neutral and Positive sentiments, with fewer misclassifications. This shows that convolutional architectures are more efficient in learning local semantic patterns in short text reviews. Taken together, these results give empirical responses to the research questions by demonstrating (1) the difference in user sentiments between ChatGPT and DeepSeek, (2) the perception of language models in terms of functionality and service quality, and (3) which deep learning model is more generalisable in terms of sentiment classification. In general, Bi-LSTM demonstrated consistent predictive ability and exemplary performance in Positive sentiment detection, whereas CNN showed better generalisation and separation of classes across all sentiment categories.

## 10 Comparative Analysis of DL Models Based on ChatGPT and DeepSeek Reviews

1. CNN performed well due to the shortness of user reviews, which are phrase-based and sentiment-rich, making them more suitable for convolutional feature extraction.
2. Bi-LSTM performed worse because it relied on sequence context, which was not as important in this dataset.
3. Both models did very well in the Positive category, but CNN had a better balance in all categories, especially reducing the errors in the Neutral and Negative sentiments.
4. This comparison shows that CNN is more efficient and accurate in user review sentiment analysis, whereas Bi-LSTM can be more applicable to more complex or longer textual data.

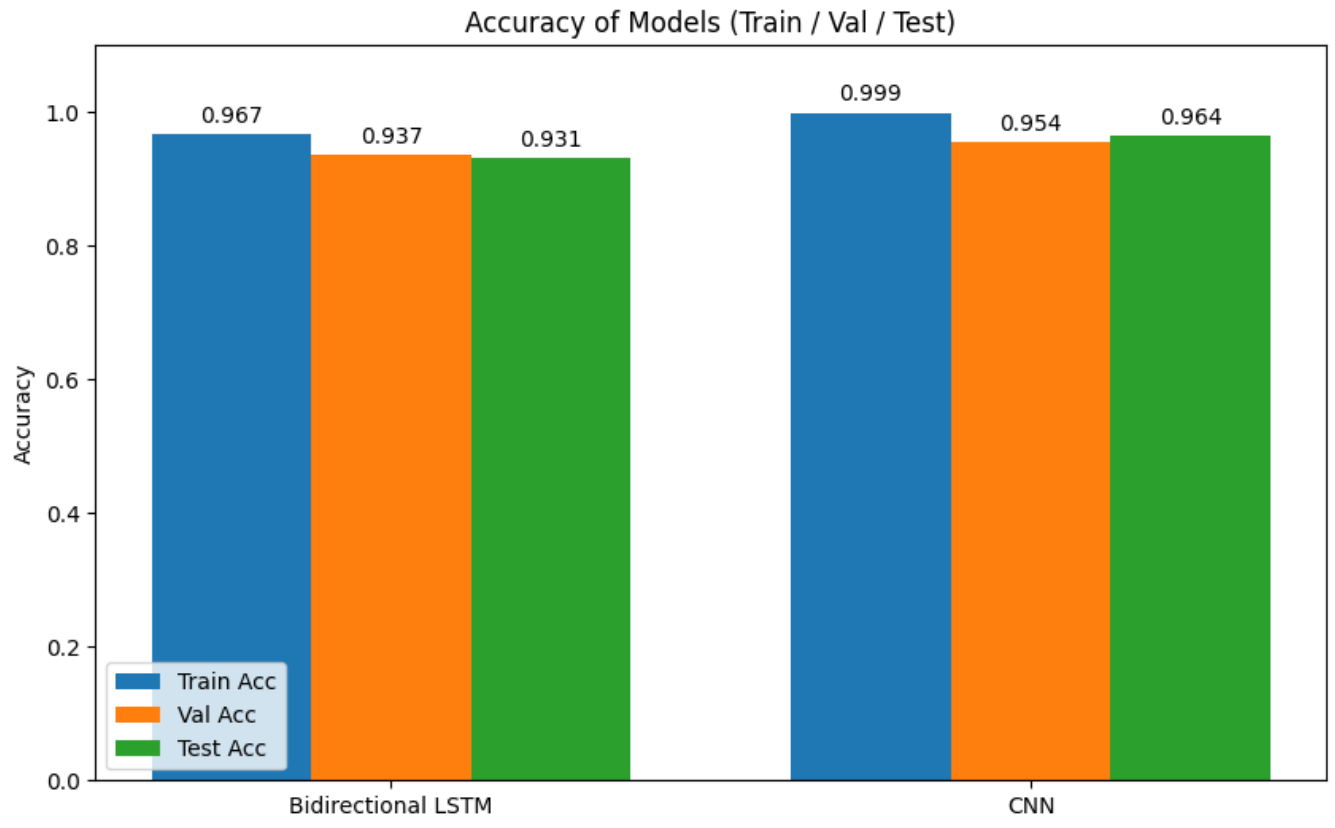

**Figure 21. Comparison of the Accuracy of Proposed DL Models**

| Work | Datasets | Method | Best of Accuracy |
|---|---|---|---|
| **Alsaleh et al. [26], 2024** | **APP review (aspect reviews)** | **Hybrid with ABSA (BERT+ Bi-LSTM + CNN)** | **94%** |
| **R. Eliviani et al. [27]. 2024** | **Google Play Reviews** | **Unspecified DL methods** | **77.7%** |
| **Samanmali et al. [28], 2025** | **Google Play reviews** | **TD-IDF with ANN, LSTM, SVM** | **85%** |
| **Y. Amirkhalili, et al. [29], 2025** | **Google Play reviews and iOS** | **NLTK + LDA + LSTM** | **82%** |
| **Our Proposed Work, 2025** | **ChatGPT and DeepSeek Google Play Store** | **CNN<br>Bi-LSTM** | **96.41%<br>93.12%** |

## Conclusion

This paper presents a comparative sentiment analysis of ChatGPT and DeepSeek, integrating two complementary perspectives: firstly, LLM-based sentiment analysis using TextBlob, which covers user perceptions, and secondly, a deep learning-based binary classification of user reviews based on a tenfold oversampled dataset of 4,000 reviews.

The findings entail a two-pronged judgment by comparing LLMs based on user-based feedback and predictive modeling performances. The empirical results demonstrate that ChatGPT is superior to DeepSeek in terms of user satisfaction, as it attracts more positive sentiment and receives higher ratings on key dimensions, including utility, responsiveness, and user experience, as of April 20, 2025. DeepSeek, in turn, had a more significant number of negative reviews, primarily due to server connectivity and technical reliability issues.

The deep learning CNN model showed the highest performance, achieving a test accuracy of 96.41% and F1-scores of 99%, 95%, and 95% for the negative, neutral, and positive classes, respectively. In comparison, the Bi-LSTM model achieved an accuracy of 93.12% with F1-scores of 99% (Negative), 90% (Neutral), and 91% (Positive). The excellent findings of CNN support its ability to extract local n-gram contents and classify the categories of sentiment acknowledged in fewer misclassifications.

## Ethics and Privacy Considerations

User review information for ChatGPT and DeepSeek was gathered from the Google Play Store through automated data scraping methods, relying exclusively on publicly accessible content. To protect user privacy, the data processing involved eliminating all personal or sensitive details that could identify individuals, and no data was utilised or retained that could lead to a privacy infringement. Legal regulations and ethical guidelines concerning the collection and application of open-source data were adhered to, with a strong emphasis on the responsible use of data solely for research purposes, without directly sharing or distributing raw data that may contain personal information.